\pdfoutput=1
\documentclass[11pt]{article}

\usepackage{acl}

\usepackage{times}
\usepackage{latexsym}
\usepackage{graphicx}
\usepackage[nolist]{acronym}
\usepackage{multirow}

\usepackage{comment}

\usepackage{lipsum}

\newcommand\blfootnote[1]{%
  \begingroup
  \renewcommand\thefootnote{}\footnote{#1}%
  \addtocounter{footnote}{-1}%
  \endgroup
}

\usepackage[T1]{fontenc}
\usepackage[utf8]{inputenc}

\usepackage{microtype}

\title{Semantic Similarity-Based Clustering of Findings \\From Security Testing Tools}

\author{Phillip Schneider$^{1*}$, Markus Voggenreiter$^{2*}$, Abdullah Gulraiz$^{1*}$ \\ {\bf and Florian Matthes$^1$} \\
         $^1$Technical University of Munich, Department of Computer Science, Germany \\
         $^2$Siemens Technology  \& LMU Munich, Germany \\ 
         \texttt{\{phillip.schneider, abdullah.gulraiz, matthes\}@tum.de} \\
         \texttt{markus.voggenreiter@siemens.com}}

\begin{acronym}
    \acro{rq}[RQ]{research question}
    \acroplural{rq}[RQs]{research questions}
    \acro{nlp}[NLP]{Natural Language Processing}
    \acro{plm}[PLM]{pretrained language model}
    \acroplural{plm}[PLMs]{pretrained language models}
    \acro{ai}[AI]{Artificial Intelligence}
    \acro{sast}[SAST]{static application security testing}
    \acro{dast}[DAST]{dynamic application security testing}
\end{acronym}

\begin{document}
\maketitle
\blfootnote{* The first three authors have contributed equally.} 
%Fast pace in DevOps and relevancy of cybersecurity
%Automated code analysis with security testing tools, resulting into duplicates
%Problem: time cost, repetitive, manual checks can lead to human error, hinderance of agility
%RQ: How do semantic similarity techniques from NLP perform in the task semantically grouping security tool reports?
%Contributions are two-fold (1) annotation tool and human-annotated corpus of deduplicated security tool reports (2) comparison different semantic similarity techniques for clustering security findings + insights from evaluation for scholars and practitioners

\begin{abstract}
Over the last years, software development in domains with high security demands transitioned from traditional methodologies to uniting modern approaches from software development and operations (\mbox{DevOps}). Key principles of \mbox{DevOps} gained more importance and are now applied to security aspects of software development, resulting in the automation of security-enhancing activities. In particular, it is common practice to use automated security testing tools that generate reports after inspecting a software artifact from multiple perspectives. However, this raises the challenge of generating duplicate security findings. 
To identify these duplicate findings manually, a security expert has to invest resources like time, effort, and knowledge. A partial automation of this process could reduce the analysis effort, encourage \mbox{DevOps} principles, and diminish the chance of human error.
In this study, we investigated the potential of applying Natural Language Processing for clustering semantically similar security findings to support the identification of problem-specific duplicate findings. 
Towards this goal, we developed a web application for annotating and assessing security testing tool reports and published a human-annotated corpus of clustered security findings. In addition, we performed a comparison of different semantic similarity techniques for automatically grouping security findings. Finally, we assess the resulting clusters using both quantitative and qualitative evaluation methods.
\end{abstract}

\section{Introduction}
The automation of security tests is a common practice for software engineering projects that apply software development and operations (\mbox{DevOps}) practices. Different security tools employ different perspectives to scan a software artifact as part of Continuous Integration or Continuous Deployment (CI/CD)  pipelines, producing semi-structured reports of security findings. While this approach fosters \mbox{DevOps} principles, reduces manual effort, and shifts security efforts to the earlier stages of development, it also comes at a cost.

Since security testing tools often have an overlapping scanning coverage, duplicates or nearly identical findings are unavoidable. Further, considering that each iteration brings new security findings, identifying duplicate security findings is essential to achieve a reliable overview. In this context, it is important to note that we define \textit{duplicates} as findings that point out the exact same security problem, potentially occurring at multiple locations in the software. Exemplary for that would be an SQL injection vulnerability at multiple locations of a web interface. Amongst multiple other activities, the identification of duplicates is traditionally addressed by a team member with security domain knowledge, a so-called security professional, before looping back the security findings to development to improve the software security-wise \cite{simpson_safecode_2014}. Taking the frequency of new reports and the number of findings throughout all security tests into account, an entirely manual analysis is unfeasible, prone to human error, and violates fundamental \mbox{DevOps} principles. 

\ac{nlp} has been shown to be effective in analyzing and clustering textual data from various application domains, such as medicine, linguistics, and software engineering \citep{demner-fushman-lin-2006-answer, majewska-etal-2018-acquiring, Aggarwal_2017}. Although security tool reports contain highly domain-specific text, it seems promising to investigate \ac{nlp} techniques for automatically grouping findings into problem-oriented clusters, which can assist security professionals in their analyses. To our best knowledge, no studies specifically focus on the machine-generated finding texts produced by security scanning tools. Addressing this research gap, we evaluated the performance of three common semantic similarity techniques. The selected techniques originate from knowledge-based, corpus-based, and neural network-based methods. Our main contributions are twofold:

\begin{enumerate}
    \item We publish a human-annotated corpus of clustered security findings along with the annotation tool used by the security professionals.
    \item We perform an in-depth analysis of three popular semantic similarity techniques for clustering security findings, followed by a quantitative and qualitative evaluation of the results.
\end{enumerate}

The remainder of this paper is structured as follows. Section~\ref{sec:related_work} presents background information on security scanning tools and gives an overview of related work on applying NLP techniques in the software engineering domain. Section~\ref{sec:method} describes the employed two-stage research approach for the dataset construction and experimental evaluation. We report the clustering results, discuss our observations, and outline the limitations in Section~\ref{sec:results}, Section~\ref{sec:discussion}, and Section~\ref{sec:limitations}, respectively. Section~\ref{sec:conclusion} concludes the paper with a summary and an outlook toward future work.

\section{Background and Related Work}
\label{sec:related_work}
This section provides background information on security testing tools and security finding reports in \mbox{DevOps}. Furthermore, we mention related studies concerning the application of \ac{nlp} techniques in the software engineering domain.

To tackle the challenge of duplicates in security reports, we first establish the definition of what duplicate security findings are. We consider two findings to be duplicates if they describe the exact same problem at any location of the software. Consequently, the same issue, e.g., an SQL injection, could occur at multiple places but would be considered a duplicate. Besides the problem-based approach, other strategies for describing duplicates can also incorporate the location of a finding or its underlying solution. The selection of a strategy in this area highly depends on the subsequent actions on the dataset. 

Furthermore, it is necessary to explain the activities that generate security reports that contain duplicate findings. Security testing can be categorized according to multiple properties depending on the testing strategy, involved testers, tested components, and numerous others.  
We limit our categorization to those security tests that can be automated in pipelines and scan an actual part of the product. Further, we categorize them into two major categories: tests that examine the static elements of the software (e.g., code, configuration, or dependencies) are called \ac{sast} and tests performed against the dynamic, actually running application are called \ac{dast}. This separation represents a clear distinction, as static testing can only guess whether a finding is actually affecting the software, while dynamic techniques directly identify the exploitable security finding. 

From our analysis of the literature on security findings management, we found that there are no \ac{nlp}-related publications that focus on the identification of duplicate security findings. However, a number of \ac{nlp} methods have been successfully applied to related subdomains in the software engineering field. For example, \citet{KUHN2007230} use latent semantic indexing (\textit{LSI}) and clustering to analyze linguistic information found in source code, such as identifier names or comments, to reveal topics and support program comprehension. In a study from \citet{Schneider_2020}, a corpus of app reviews with comments about a variety of software issues is clustered into topics with problem-specific issue categories. Another study from \citet{info9090222} focuses on automatically forming semantic clusters of functional requirements based on cosine similarity with a corpus of documents containing software requirements specifications. The authors conduct an empirical evaluation of agglomerative hierarchical clustering using four open-access software projects. In order to assess the software quality of programs, \citet{6079848} apply a hierarchical cluster algorithm to create problem-oriented clusters, reducing the effort needed to review the code. The study shows that semantic clusters are an effective technique for defect prediction.

\section{Method}
\label{sec:method}
%To achieve our objective of investigating semantic similarity techniques for clustering findings from security testing reports, we constructed a human-annotated dataset with 1351 \ac{sast} and 36 \ac{dast} findings. The two-stage process of dataset construction and experimental evaluation is explained in the next subsections.
In order to achieve our objective of investigating semantic similarity techniques for clustering findings from security testing reports, we constructed a human-annotated dataset. This annotated corpus consists of 1351 \ac{sast} and 36 \ac{dast} findings. The two-stage process with dataset construction as well as experimental evaluation is explained in the following subsections.

\subsection{Dataset Construction}
To quantify the performance of different semantic similarity techniques, a ground-truth benchmark dataset is required, enabling the comparison between human-labeled clusters and the predictions of the semantic similarity algorithms. Therefore, we asked two security professionals from the industry to annotate semantically duplicate findings in a given list of security reports. Due to the significant differences in perspective between \ac{sast} and \ac{dast} reports, we decided to construct two separate datasets, each of which comprising reports from only one testing type.

A major challenge in constructing such a dataset is the content of the security tool reports. Security tool reports are often exported as JSON files containing security finding objects. Across different tools, these reports utilize different schemas, resulting in different property names referring to the same finding feature (e.g., \textit{description}, \textit{FullDescription}, \textit{text}, \textit{Message}, or \textit{details}). For the construction, the security professionals consolidate semantically duplicate findings from all tool reports of a testing iteration based on certain features, e.g., description, location, or unique identifier. Therefore, they need to find the feature in the respective tool schema and compare it to the other findings. Manually annotating such a dataset would require them to memorize \(N \times M\) property names when identifying \(N\) features across \(M\) distinct security testing reports. 
To enhance efficiency and reduce manual, repetitive work, we developed the Security Findings Labeler (\textit{SeFiLa}).\footnote{https://github.com/abdullahgulraiz/SeFiLa} This tool allows security professionals to upload reports from different security tools and conveniently group all findings into named clusters. 

The initial, unconsolidated reports of the dataset were generated by scanning the open-source, vulnerable web application JuiceShop\footnote{https://owasp.org/www-project-juice-shop/} with seven \ac{sast} tools and two \ac{dast} tools. For reproducibility reasons, we solely selected tools free of charge that can be reasonably automated in real-world software development pipelines. We selected Anchore, Dependency Check, Trivy, HorusSec, Semgrep, CodeQL, and Gitleaks as \ac{sast} tools. For \ac{dast}, we selected Arachni and OWASP ZAP. Fundamental information about each tool can be found in Table~\ref{tab:tool_overview} in the appendix. From each tool, one testing report was taken for the dataset. The security professionals assigned findings to named clusters representing the same security problem. This process was aided by features like the CVE-ID (common vulnerabilities and exposures) which provides an identifier and a reference-method for publicly known security vulnerabilities. Other helpful features are descriptions and solutions generated by the testing tools. After all findings were assigned to clusters, the dataset comprising our baseline for duplicate identification was completed. The dataset and the code to run the test cases were published in a public GitHub repository.\footnote{https://github.com/abdullahgulraiz/SeFiDeF}

\subsection{Evaluation Procedure}
For conducting the evaluation, we investigated semantic similarity methods proposed in the literature and chose three popular techniques that are often used as baseline models: knowledge graph-based similarity with \textit{WordNet} \cite{miller_wordnet}, \textit{LSI} \cite{landauer1997AST}, and \textit{SBERT} \cite{reimers-2019-sentence-bert}. To evaluate the semantic similarity techniques, we extracted all findings from the security testing tool reports and concatenated selected features from them to form problem-specific finding strings. We applied the three chosen semantic similarity techniques to the finding strings to determine those that are semantically similar. Since semantic similarity between two finding strings is calculated as a score between 0 and 1 where 1 indicates highest similarity, we established a \textit{similarity threshold} for each experiment. This threshold defines the value above which two finding strings are deemed to be semantically similar. Findings corresponding to these similar finding strings are then grouped to form predicted clusters. Implementation-wise, predicted and ground-truth clusters both consist of unique integer sequences, each integer representing a finding from the dataset. 

Before the clusters were compared with each other in the quantitative evaluation, we encountered the need for \textit{transitive clustering} of findings. In certain cases, the problem description of two findings was identical, but it was repeated in one finding for multiple instances, leading to a discrepancy in text length. Since the similarity depends on the similarity of the finding strings, we encounter the following example predictions with \textit{Similar Findings} listed in descending order of semantic similarity scores with the corresponding \textit{Finding} identifier:
\pagebreak
\[
\{Finding: 1,\:Similar\:Findings: \{1, 2, 4\}\}
\]
\[
\{Finding: 2,\:Similar\:Findings: \{2, 1, 3, 5\}\}
\]

 Let us assume that findings \(\{1, 2, 3\}\) contain the same problem description, although it appears once in \(Finding\:1\), two times in \(Finding\:2\), and three times in \(Finding\:3\). While \(Finding\:1\) is found similar to findings \(\{1, 2, 4\}\), its similarity score with respect to \(Finding\:3\) is below the clustering threshold due to the different text length. However, \(Finding\:2\) does have \(Finding\:3\) in its set of similar findings. If \(Finding\:3\) is similar to \(Finding\:2\), it should also be similar to \(Finding\:1\), regardless of repetitive text. Therefore, even though \(Finding\:3\) exists only in the set of similar findings for \(Finding\:2\), it should appear in the final set of similar findings of \(Finding\:1\) as well. In our initial clustering experiments and discussions with the security professional, we observed that while lowering the similarity threshold led to many false positive predictions, transitive clustering improved the results without changing the similarity threshold. Therefore, we apply the transitive property to consider findings as semantically related through \textit{intermediate} findings. This causes the above predictions to become: 
 
\[
\{Finding: 1,\:Similar\:Findings: \{1, 2, 3, 4, 5\}\}
\]
 \[
\{Finding: 2,\:Similar\:Findings: \{1, 2, 3, 4, 5\}\}
\]

After transitive clustering, we removed the duplicate clusters from predictions and evaluated the final predictions against the ground-truth clusters.

Table~\ref{tab:contingency_matrix} shows a contingency matrix that illustrates possible outcomes when comparing clusters from predictions (P) with clusters from the ground-truth dataset (Q). The number of occurrences of these outcomes is used to calculate the metrics of \textit{precision}, \textit{recall}, and \textit{F-score}.

\begin{table}[htpb]
  \centering
  \small
  \begin{tabular}{p{2.1cm}p{1.9cm}p{2.1cm}}
    \hline
      & \multicolumn{2}{c}{\textbf{Predictions (P)}} \\
      & Clusters in P & Clusters not in P\\
      \hline
    \textbf{Ground-truth (Q)} & & \\
      Clusters in Q & True Positive (TP) & False Negative (FN) \\
      Clusters not in Q & False Positive (FP) & True Negative (TN) \\
    \hline
  \end{tabular}
  \caption{Contingency matrix of predicted clusters P and ground-truth clusters Q.}
  \label{tab:contingency_matrix}
\end{table}

The \textit{precision}~\cite{hossin2015review} measures positive patterns correctly predicted from the total predicted patterns in a positive class. In our experiments, it measures the ratio of correct cluster predictions to all predictions. Higher precision indicates that less false positive predictions appeared in the results. It is calculated as:
\[
Precision = \frac{TP}{TP+FP}
\]

The \textit{recall}~\cite{hossin2015review} is used to measure the fraction of correctly classified positive patterns. In our experiments, it represents the ratio of correctly predicted clusters to all ground-truth clusters. A high recall value thus indicates that the semantic clustering results retrieve many ground-truth clusters of the security professional.
\[
Recall = \frac{TP}{TP+FN}
\]

The \textit{F-score}, also known as Dice Measure~\cite{dice1945measures}, calculates the harmonic mean between precision and recall. It balances both metrics to provide an overall performance overview.
\[
F-score = \frac{2*TP}{2*TP+FP+FN}
\]

In addition to the quantitative evaluation with performance measures, we collected qualitative feedback from the security professionals on incorrectly clustered findings. We limited the information about each finding to the finding strings used as input for the \ac{nlp} techniques and asked for possible reasons for the incorrect clustering. This created a list of reasons that led to poor duplicate identification from the perspective of a domain-aware security professional. Finally, each incorrect cluster is associated with at least one reason for the incorrect clustering, providing insights into the different challenges and their prevalence in the results. The evaluation was aided by \textit{SeFiLa} for annotation of the findings, assignment of reasons, and documentation.

\section{Experiments}
\label{sec:results}
\subsection{Dataset Description}
After labeling the exported security findings with our annotation tool \textit{SeFiLa}, the security professionals provided us with two datasets, namely the manually grouped \ac{sast} and \ac{dast} findings. The descriptive statistics of both datasets are summarized in Table~\ref{tab:dataset_summary}. We observe that \ac{sast} findings are far more frequent, making up 97.4\% of the total findings. The number of formed clusters for the \ac{sast} findings is significantly higher than for \ac{dast} findings. While both datasets had clusters with only one finding, the maximum cluster size was by far larger in the \ac{sast} dataset. Despite these discrepancies, the average number of findings per cluster is not too different between the datasets, ranging from a mean value of 3 for \ac{dast} to a mean value of 7 for \ac{sast} findings. In addition, \ac{dast} finding texts are more verbose since they contain 169 more characters on average.
\begin{table}
    \centering
    \begin{tabular}{p{4.25cm}cc}
        \hline
        \textbf{Statistic} & \textbf{SAST} & \textbf{DAST} \\
        \hline
        Number of clusters & 183 & 10 \\
        Number of findings & 1351 & 36 \\
        \hline
        Avg. findings per cluster & 7 & 3 \\
        Avg. characters per finding & 302 & 471 \\
        \hline
        Min. findings per cluster & 1 & 1 \\
        Max. findings per cluster & 408 & 25 \\
        \hline
    \end{tabular}
    \caption{Data records from static analysis security tools (SAST) and dynamic analysis security tools (DAST).}
    \label{tab:dataset_summary}
\end{table}
To investigate the potential of semantic similarity techniques, constructing the finding string from the finding features is crucial. 
Analyzing the initial dataset, we identified that solely a single feature describing the finding is consistently found across all \ac{sast} tools. For the \ac{dast} findings, multiple features, including the description, a name, and even a solution/mitigation, were consistently found across all findings. Furthermore, we observed that \ac{dast} features are sufficiently verbose to comprehend the problem from their finding string and thereby contain enough semantic content for \ac{nlp}. Contrarily, we find \ac{sast} features to be very brief, for that matter, making it almost impossible to understand a finding just from the finding string.  

To counteract the limitation of very short \ac{sast} finding strings, we make use of CVE-IDs to increase the textual content of \ac{sast} finding strings. By leveraging the CVE identifier present in some findings, we concatenated finding strings of various machine-generated descriptions with the same CVE-ID. This allows for more semantic content and longer descriptions about the underlying problem. We used the concatenated finding strings as input to the \ac{nlp}-based similarity techniques. 

This step led us to construct a total of four corpora with finding strings from both \ac{sast} and \ac{dast} datasets for the identification of duplicate findings, as listed below:
\begin{itemize}
\itemsep-0.3em
  \item \textbf{\textit{SAST-D}:} consists only of SAST finding descriptions
  \item \textbf{\textit{SAST-ConcD}:} consists of concatenated SAST finding descriptions with the same CVE-ID
  \item \textbf{\textit{DAST-NDS}:} consists of concatenated DAST finding names, descriptions, and solution texts
  \item \textbf{\textit{DAST-D}:} consists only of DAST finding descriptions
\end{itemize}

\subsection{Evaluation Results}

% -------------- FOR CORRECT PLACEMENT, HAS NO USE HERE -----------------
\begin{figure*}[htpb]
  \centering
  \includegraphics[width=\linewidth]{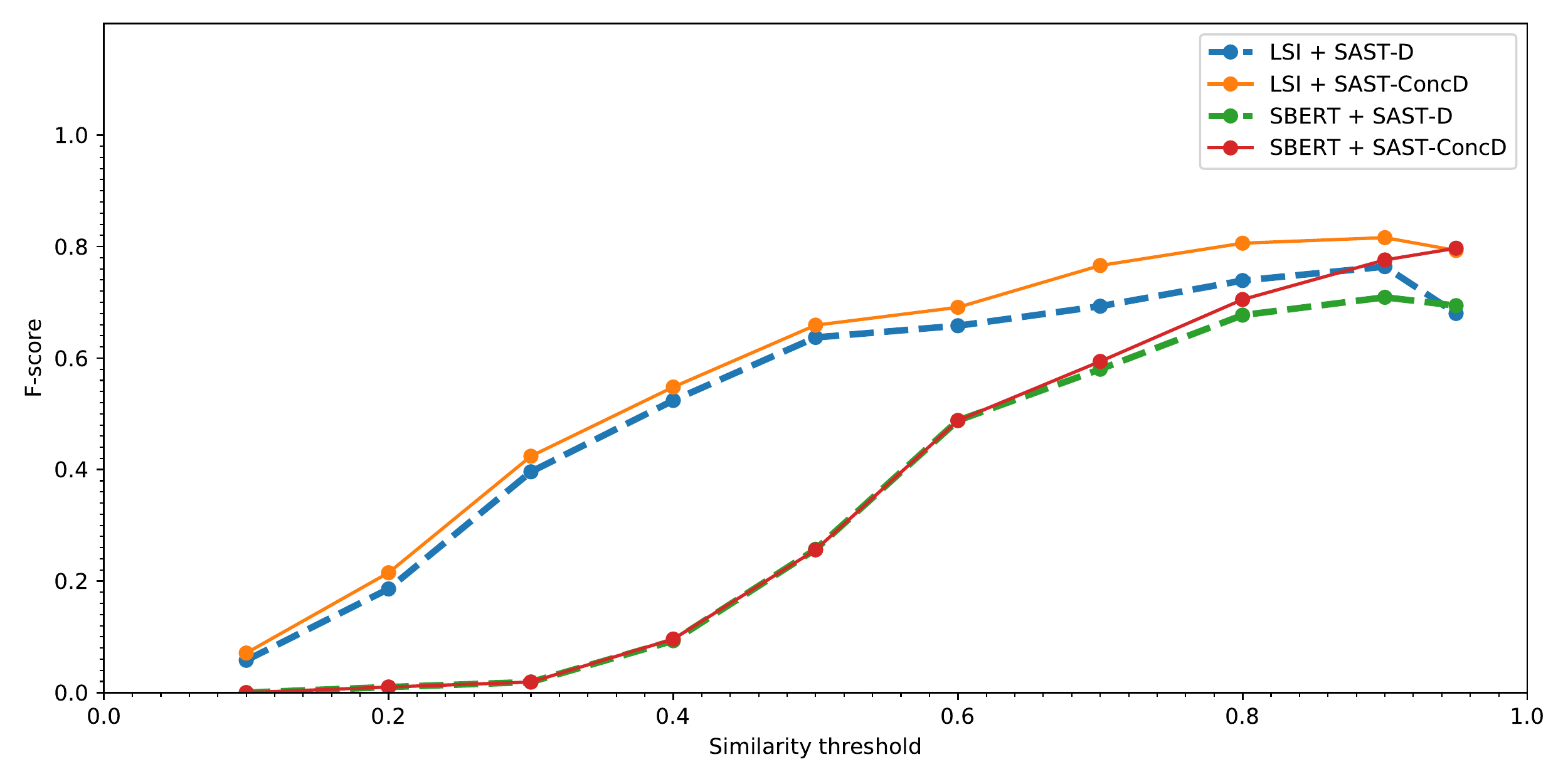}
  \caption{Semantic clustering results of SAST findings for different similarity thresholds.}\label{fig:sast_results_plot}
\end{figure*}
% -------------- -----------------

The summary of the quantitative results achieved when applying semantic clustering using a technique from each category of semantic similarity methods to each of the four corpora is presented in Table~\ref{tab:summary_quantitative_results}. The experiments were performed for similarity thresholds \( 0.1 \le \) and \(\le 0.95 \). The performance metric values for the experiment with the highest F-score are reported.

% report table with overview of performance metrics
\begin{table}[htpb]
  \centering
  \small
  \begin{tabular}{l p{1.2cm} c c c c}
    \hline
      \multirow{2}{*}{Technique} & \multirow{2}{*}{Corpus} & \multicolumn{3}{c}{Metrics} \\
      & & F-score & Precision & Recall \\
    \hline
    \multirow{4}{*}{SBERT} & \textit{SAST-D} & 0.709 & 0.621 & 0.825 \\
    & \textit{SAST-ConcD} & 0.797 & 0.701 & 0.923 \\
    & \textbf{\textit{DAST-NDS}} & \textbf{0.857} & \textbf{0.818} & \textbf{0.900} \\
    & \textit{DAST-D} & 0.857 & 0.818 & 0.900 \\
    \hline
    \multirow{4}{*}{LSI} & \textit{SAST-D} & 0.739 & 0.658 & 0.842 \\
    & \textbf{\textit{SAST-ConcD}} & \textbf{0.816} & \textbf{0.734} & \textbf{0.918} \\
    & \textit{DAST-NDS} & 0.857 & 0.818 & 0.900 \\
    & \textit{DAST-D} & 0.857 & 0.818 & 0.900 \\
    \hline
    \multirow{4}{*}{KG} & \textit{SAST-D} & 0.659 & 0.556 & 0.809 \\
    & \textit{SAST-ConcD} & 0.777 & 0.676 & 0.913 \\
    & \textit{DAST-NDS} & 0.727 & 0.667 & 0.800 \\
    & \textit{DAST-D} & 0.727 & 0.667 & 0.800 \\
    \hline
  \end{tabular}
  \caption{Summary table of performance metrics (highlighted results show the best performing techniques for SAST and DAST).}
  \label{tab:summary_quantitative_results}
\end{table}
 
\subsubsection{Comparison of Semantic Similarity Techniques}
Figure~\ref{fig:sast_results_plot} and Figure~\ref{fig:dast_results_plot} show the F-scores of different technique-corpus combinations over different similarity thresholds for \ac{sast} and \ac{dast}, respectively. We see that the F-scores increase with increasing similarity threshold, peaking at a threshold value \( \ge 0.6 \) for DAST and at around 0.9 for SAST. Figure~\ref{fig:kg_results_plot} in the appendix shows the performance metrics for clustering with knowledge graph-based semantic similarity. It is noteworthy that the F-scores for the knowledge graph-based clustering are not only lower in comparison to \textit{LSI} and \textit{SBERT} but they also reach a plateau for threshold values higher than 0.2.

% discussion of performance metrics of individual, describe effect of threshold parameter and dataset
%The summary of the quantitative results achieved when applying semantic clustering using a technique from each category of semantic similarity methods to each of the four corpora is depicted in Table~\ref{tab:summary_quantitative_results}. The experiments were performed for \( 0.1 \le similarity\:threshold \le 0.95 \), and the metric values for experiment with \textit{highest F-score} are reported. 

%Figure~\ref{fig:sast_results_plot} and Figure~\ref{fig:dast_results_plot} show the F-scores of different technique-corpus combinations over different similarity thresholds for \ac{sast} and \ac{dast}, respectively. It can be seen that the F-scores increase with increasing similarity threshold, and that the highest value for each combination is achieved at \( similarity\:threshold \ge 0.6 \).

\subsubsection{Qualitative Evaluation}
For the qualitative evaluation, we showed incorrect predictions from the best results of semantic clustering of \ac{sast} and \ac{dast} findings to a security professional. The cluster results came from applying \textit{LSI} to \textit{SAST-ConcD} corpus for the \ac{sast} dataset and applying \textit{SBERT} to \textit{DAST-NDS} corpus for the \ac{dast} dataset. Using \textit{SeFiLa}, the security professional inspected incorrect predictions and their associated ground-truth cluster. The security professional assigned possible reasons for poor duplicate identification by reading finding strings associated with incorrect predictions. These reasons are documented for 72 incorrect \ac{sast} predictions and 2 incorrect \ac{dast} predictions. The reasons and the number of times they were assigned to an incorrect prediction from either \ac{sast} or \ac{dast} clusters are listed in Table~\ref{tab:summary_qualitative_results}. 
%For qualitative evaluation, we provide incorrect predictions from the best results of semantic clustering of both \ac{sast} and \ac{dast} findings to our security professional. The results come from applying \textit{LSI} to \textit{SAST-ConcD} corpus for the \ac{sast} dataset, and applying \textit{SBERT} to \textit{DAST-NDS} corpus for the \ac{dast} dataset. Using \textit{SeFiLa}, the security professional viewed incorrect predictions and their associated ground-truth cluster. By reading finding strings associated with incorrect predictions, possible reasons that lead to poor duplicate identification are assigned. These reasons are attached to 72 incorrect predictions for \ac{sast} and 2 incorrect predictions for \ac{dast}, respectively. The list of reasons along with the number of times they were assigned to an incorrect prediction from either \ac{sast} or \ac{dast} clusters is outlined in Table~\ref{tab:summary_qualitative_results}. 

\section{Discussion}
\label{sec:discussion}
% 1-2 columns
% discuss key findings from quant. and qual. evaluation
% what techniques are recommended for usage

% -------------- FOR CORRECT PLACEMENT, HAS NO USE HERE -----------------
\begin{figure*}[h]
  \centering
  \includegraphics[width=\linewidth]{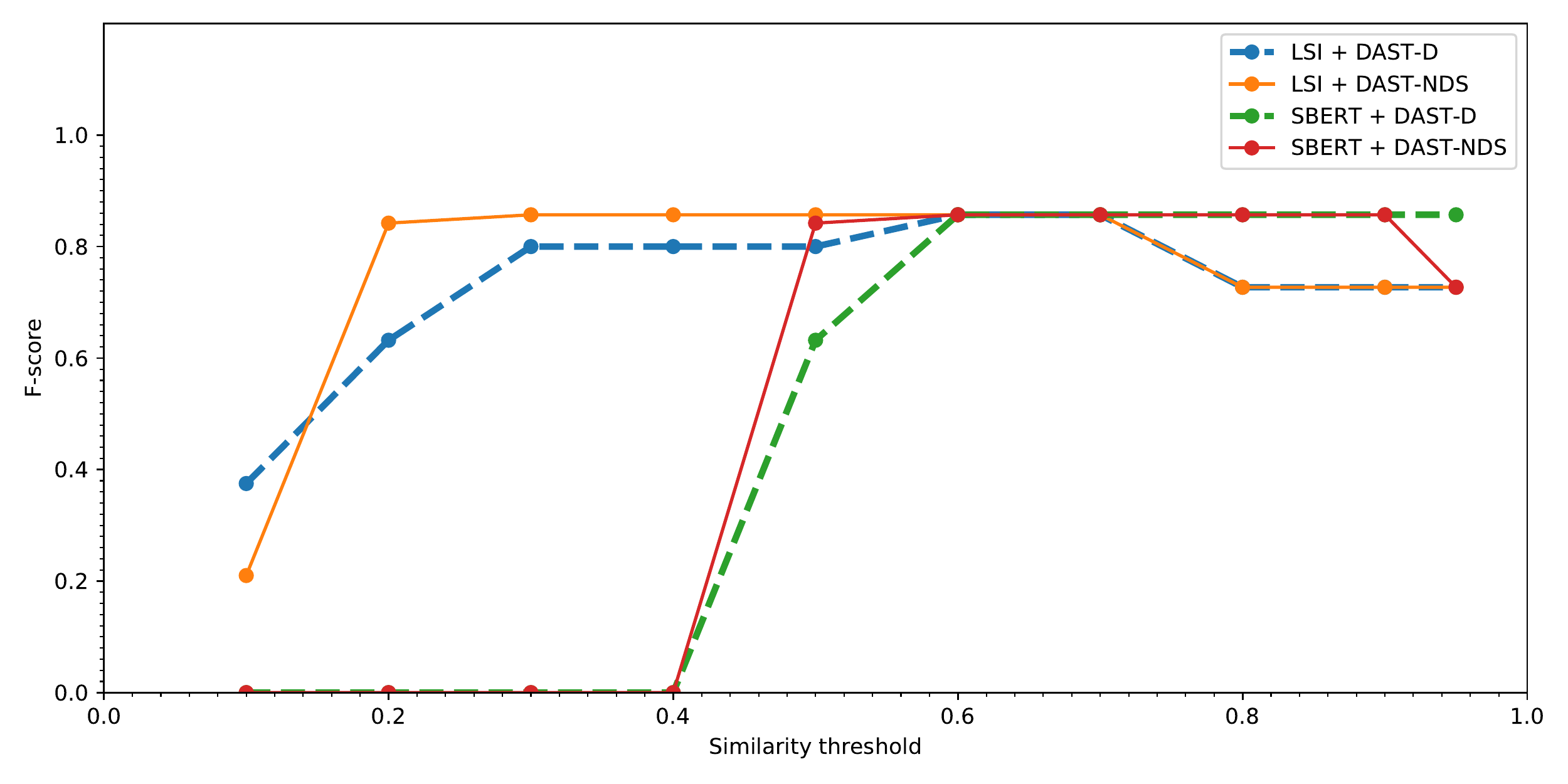}
  \caption{Semantic clustering results of DAST findings for different similarity thresholds.}\label{fig:dast_results_plot}
\end{figure*}
% -------------- -----------------
From the quantitative evaluation, we see that \ac{sast} findings are best clustered by applying \textit{LSI} to the \textit{SAST-ConcD} corpus, which gives a F-score of 0.816. Although applying \textit{SBERT} to the same corpora provides a similar F-score of 0.797 and matches a higher ratio of ground-truth clusters due to higher recall, \textit{LSI} has a higher precision and less false positive predictions, which is a crucial requirement to the security professionals. Hence, applying \textit{LSI} to \textit{SAST-ConcD} corpus is our recommendation for identifying duplicate \ac{sast} findings. 

When clustering \ac{dast} findings, we see that the highest F-score of 0.857 is achieved by applying \textit{SBERT} and \textit{LSI} to both \textit{DAST-D} and \textit{DAST-NDS} corpora. However, as illustrated in Figure~\ref{fig:dast_results_plot}, applying \textit{SBERT} yields a high F-score for similarity threshold \(\ge 0.6 \), whereas \textit{LSI} yields a lower F-score. Since higher similarity thresholds are preferred in production scenarios to prevent false positive predictions, \textit{SBERT} is preferred over \textit{LSI}. For the corpus, \textit{DAST-NDS} is preferred over \textit{DAST-D} due to more textual content from three features, which leads to a better grasping of semantics and provides better distinction amongst false positives. We also see that for similarity threshold \( > 0.9 \), the F-score of \textit{SBERT} with \textit{DAST-NDS} slightly decreases. This is because of the strict distinction by semantic similarity algorithms, which also consider the semantics of a problem's solution when distinguishing between problems identified by different findings.

From the qualitative evaluation, we see that a significant challenge for \ac{sast} findings is the content of the finding description. Some tools provide a title instead of an actual description of the underlying problem. This leads to insufficient semantic content being derived from the finding corpus texts, thereby leading to poor duplicate identification. Another frequent reason for incorrect predictions in \ac{sast} are suboptimally constructed finding strings. This primarily arises when the information content of the original finding is low, which even is challenging for security professionals when determining duplicate findings just from reading the findings string. The third most highlighted challenge for \ac{sast} is the imbalance in the verbosity of description features of findings. The description either contains contextually rich problem descriptions or under-specified ones, leading to different extents of semantic content being captured and, thereby, incorrect comparisons being made. For \ac{dast}, we only have two incorrect predictions, which can be traced back to the challenge of being very application- and domain-specific.

To improve \ac{sast} findings clustering results, it is evident that the semantic content of finding strings representing a problem must be improved. This can be done by using multiple external sources, e.g., the National Vulnerability Database\footnote{https://nvd.nist.gov/}, the GitHub Advisory Database\footnote{https://github.com/advisories} for enrichment, or by scraping information from the multiple reference sources listed in a finding. The goal is that the textual data of each finding consists of multiple paragraphs and contains enough semantic content for the semantic similarity techniques to grasp as much contextual information as possible. Furthermore, the final corpus texts should contain the same verbosity level to avoid a bias related to the text length. Lastly, the same clustering approach can be studied using \ac{nlp} models that are fine-tuned for security findings, accounting for the domain-specific vocabulary to improve the clustering results. 

\begin{table*}[htpb]
    \centering
    \small
    \begin{tabular}{cp{0.7\linewidth}cc}
        \hline
        \textbf{Reason} & \textbf{Explanation for Incorrect Clustering} & \textbf{SAST} & \textbf{DAST} \\
        \hline
        1 & In the context of the product, this result can only be identified by somebody knowing the context of the application. & - & 2 \\
        2 & Different tools use a different phrasing to explain the same issue. & 5 & - \\
        3 & The tools sometimes provide no description of the finding. Hence, the features could only rely on the title. & 39 & - \\ 
        4 & Some tools provide more and some tools provide less text in their description, which reduces the impact of actual relevant features. & 19 & - \\
        5 & Additional review necessary due to an unknown reason for the decision. & 5 & - \\
        6 & The sub-optimally constructed feature string could be the reason for the incorrect clustering. & 39 & - \\
        7 & The tool describes the finding precisely according to the location of occurrence. Hence the finding text is over-specified. & 3 & - \\
        8 & Human annotation error and the suggested clustering by the algorithm is correct. & 1 & - \\
        9 & One tool addresses the issue of using an \textit{eval} function, while the other one has the problem of user controlled values in it. However, it would not be considered as a major false positive. & 3 & - \\
        \hline
    \end{tabular}
    \caption{Overview of provided explanations from the qualitative evaluation.}
    \label{tab:summary_qualitative_results}
\end{table*}
% -------------- -----------------

\section{Limitations}
\label{sec:limitations}
While we present a variety of results regarding semantic clustering of security findings, our conclusions are limited in certain aspects. Firstly, all our findings result from scanning a single web application: JuiceShop. While it contains vulnerabilities encountered in real-world applications, it is restricted in its representation of a real scenario because JuiceShop is intended to comprise multiple vulnerabilities. Moreover, the subset of JuiceShop vulnerabilities that are clustered poorly might appear most often in reality, threatening the external validity of the results. 
Furthermore, our findings result from a finite number of modern security tools. While these tools are open-source and currently widely used, the scanning functionality of security testing tools is constantly evolving. Thereby, the scanning tools we use might change based on the needs of the domain. 
Lastly, our datasets were labeled by two security professionals and the results were evaluated by one security professional. While this is beneficial to prevent inconsistencies due to the subjective nature of the annotation tasks, the relevance of our results is highly dependent on the created ground-truth dataset. However, our chosen research design aims at making the results of our work as objective as possible. Researchers and practitioners can also use our developed annotation tool to reproduce our data collection or transfer our study insights to a setting of their own choice.

\section{Conclusions and Future Work}
\label{sec:conclusion}
% What did we do
% What where the results
% How can future researchers build up on results
% How can results help practice 

In this work, we explored the applicability of semantic clustering of security findings through various similarity techniques. We tested three techniques from neural network-based, corpus-based, and knowledge-based methods on finding strings that describe security vulnerabilities identified by testing tools.

To this end, we created a ground-truth dataset of security findings clustered according to the expertise of security professionals. We compared this dataset to the results of semantic similarity techniques, indicating that \ac{sast} findings are best clustered by applying \textit{LSI} to \textit{SAST-ConcD} corpus, whereas \ac{dast} findings are best clustered by applying \textit{SBERT} to \textit{DAST-NDS} corpus. Conducting a qualitative evaluation with a security professional, we additionally pointed out the challenges encountered by semantic similarity techniques when applied to security findings and discussed possible solution strategies. 

One potential future work would be the application of the chosen techniques to cluster security findings according to other testing strategies like solution-based clustering. This could grant deeper insights into the challenges of grouping security findings with \ac{nlp} and provide access to new use cases. 
Furthermore, research on how plain neural networks perform when trained directly on semi-structured security findings appears to be promising given modern advancements in neural network architectures. Especially when compared to the \ac{nlp}-based approach in this work, the properties of neural networks are worth exploring. Since neural networks automatically prioritize important features with layers like max-pooling, the manual effort undertaken to determine problem-describing features and clustering based on them could be alleviated. However, training a neural network requires significantly more data, so the construction of a much larger findings dataset would be necessary.
Finally, an evaluation of the identified techniques in real-world \mbox{DevOps} scenarios could provide valuable insights into the practical usefulness of our approach in software development projects.

\section*{Acknowledgements}
The authors want to thank the industry professionals for their exceptional effort in clustering the security findings and evaluating the shortcomings. 

\bibliography{anthology,custom}
\bibliographystyle{acl_natbib}

\appendix
\onecolumn
\section{Supplementary Material }
\label{sec:appendix}
\begin{figure*}[h]
  \centering
  \includegraphics[width=\textwidth]{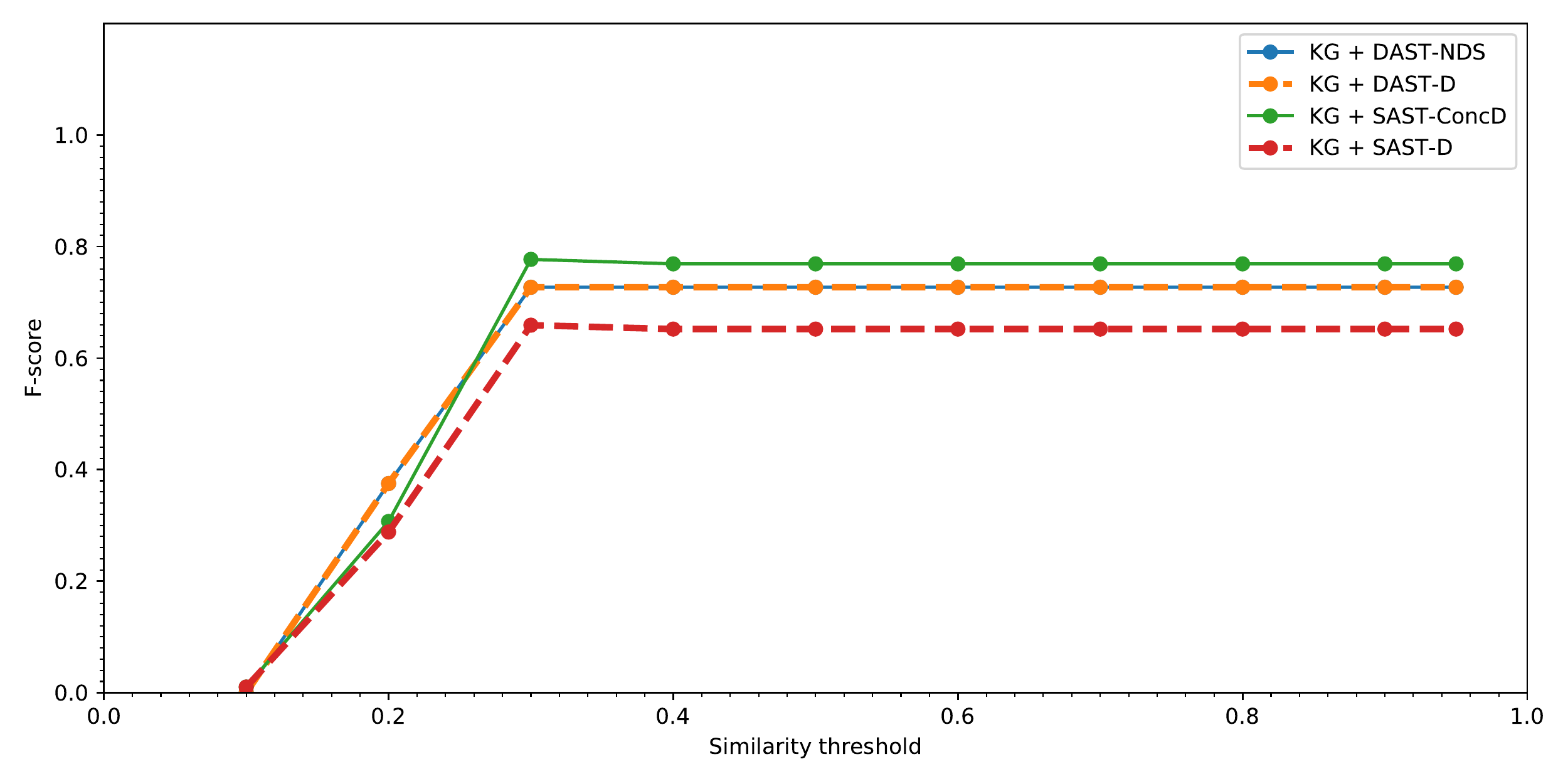}
  \caption{Semantic clustering results with knowledge graph-based similarity.}
  \label{fig:kg_results_plot}
\end{figure*}

In this appendix, we provide additional material to the main article. Table~\ref{tab:tool_overview} lists the security testing tools that were used to scan the web application JuiceShop and generate security findings. Figure~\ref{fig:kg_results_plot} shows the performance metrics for clustering with knowledge graph-based semantic similarity.

\begin{table*}[h]
\centering
    \begin{tabular}{llll}
        \hline
        \textbf{Tool} & \textbf{Category} & \textbf{Analysis Type} & \textbf{Link} \\
        \hline
        Anchore & SAST & Third-party vulnerabilities & \href{https://anchore.com/opensource/}{anchore.com/opensource} \\
        Dependency Checker & SAST & Third-party vulnerabilities & \href{https://owasp.org/www-project-dependency-check/}{owasp.org/dependency-check} \\ 
        Trivy & SAST & Third-party vulnerabilities &\href{https://github.com/aquasecurity/trivy}{github.com/aquasecurity/trivy} \\
        GitLeaks & SAST & Hardcoded secrets &\href{https://github.com/zricethezav/gitleaks}{github.com/zricethezav/gitleaks} \\
        CodeQL & SAST & Coding flaws & \href{https://codeql.github.com/}{codeql.github.com} \\
        Horusec & SAST & Coding flaws & \href{https://horusec.io/site/}{horusec.io/site} \\
        Semgrep & SAST & Coding flaws & \href{https://semgrep.dev/}{semgrep.dev} \\
        \hline
        Arachni & DAST & Web app scan & \href{https://github.com/Arachni/arachni}{github.com/Arachni/arachni} \\
        ZAP & DAST & Web app scan &\href{https://www.zaproxy.org/}{www.zaproxy.org} \\
        \hline
    \end{tabular}
    \caption{Overview of static (SAST) and dynamic (DAST) analysis security tools that were used to scan JuiceShop.}
    \label{tab:tool_overview}
\end{table*}

\end{document}